\documentclass[preprint,12pt]{elsarticle}
\graphicspath{ {./figures/} }
\usepackage{hyperref}
\usepackage{float}
\usepackage{verbatim} 
\usepackage{apalike}
\restylefloat{figure}
\floatstyle{plaintop} 
\restylefloat{table}
\usepackage{multirow}

\journal{Expert Systems with Applications}

\biboptions{authoryear}

\begin{document}
\begin{frontmatter}


\begin{titlepage}
\begin{center}
\vspace*{1cm}

\textbf{ \large Text2Graph VPR: A Text-to-Graph Expert System for Explainable Place Recognition in Changing Environments }

\vspace{1.5cm}

Saeideh Yousefzadeh$^{a}$ ( sa.yousefzadehmaghani@mail.um.ac.ir),\\ Hamidreza Pourreza$^a$ ( hpourreza@um.ac.ir) 

\hspace{10pt}

\begin{flushleft}
\small  
$^a$ Machine Vision Lab, Ferdowsi University of Mashhad, Iran \\

\vspace{1cm}
\textbf{Corresponding author at: } \\
Faculty of Engineering, Ferdowsi University of Mashhad, Azadi Sqr., Mashhad, Iran \\
Tel: +985138805025  \\
Email: hpourreza@um.ac.ir

\end{flushleft}        
\end{center}
\end{titlepage}

\title{Text2Graph VPR: A Text-to-Graph Expert System for Explainable Place Recognition in Changing Environments}

\author[label1]{Saeideh Yousefzadeh}
\ead{sa.yousefzadehmaghani@mail.um.ac.ir}

\author[label1]{Hamidreza Pourreza\corref{cor1}}
\ead{hpourreza@um.ac.ir}

\cortext[cor1]{Corresponding author.}
\address[label1]{Machine Vision Lab, Ferdowsi University of Mashhad, Iran}

\begin{abstract}
Visual Place Recognition (VPR) in long-term deployment requires reasoning beyond pixel similarity: systems must make transparent, interpretable decisions that remain robust under lighting, weather and seasonal change. We present Text2Graph VPR, an explainable semantic localization system that converts image sequences into textual scene descriptions, parses those descriptions into structured scene graphs, and reasons over the resulting graphs to identify places. Scene graphs capture objects, attributes and pairwise relations; we aggregate per-frame graphs into a compact place representation and perform retrieval with a dual-similarity mechanism that fuses learned Graph Attention Network (GAT) embeddings and a Shortest-Path (SP) kernel for structural matching. This hybrid design enables both learned semantic matching and topology-aware comparison, and—critically— produces human-readable intermediate representations that support diagnostic analysis and improve transparency in the decision process. We validate the system on Oxford RobotCar and MSLS (Amman/San Francisco) benchmarks and demonstrate robust retrieval under severe appearance shifts, along with zero-shot operation using human textual queries. The results illustrate that semantic, graph-based reasoning is a viable and interpretable alternative for place recognition, particularly suited to safety-sensitive and resource-constrained settings.
\end{abstract}

\end{frontmatter}
\section{Introduction}
\label{introduction}
Visual Place Recognition (VPR) is a fundamental component of robotics and autonomous navigation systems, where it aims to determine the location of a query image or image sequence by matching it to previously visited places. More broadly, VPR can be viewed as a decision-making problem in which an intelligent system reasons over observations to infer spatial context. This task becomes especially challenging under significant appearance variations, such as those caused by changes in lighting, weather, seasons, or viewpoint \citep{Lowry2016VPRSurvey,Garg2018Lost}.
Over the past decade, Visual Place Recognition (VPR) has seen major progress through image-based deep learning methods. Early approaches such as NetVLAD \citep{Arandjelovic2016NetVLAD} and its extensions \citep{Hausler2021PatchNetVLAD} established strong global image descriptors, while more recent sequence-based methods—including SeqNet \citep{Garg2021SeqNet}, DeepSeqSLAM \citep{Chancan2020DeepSeqSLAM}, JVPR \citep{Li2023JVPR}, and JIST \citep{Berton2023JIST}—exploit temporal continuity for improved robustness. Models like SeqMatchNet \citep{Garg2022SeqMatchNet} and TCGAT \citep{Wang2024MultimodalVPR} further enhance sequence modeling through contrastive learning and graph-based fusion of multimodal cues. While these techniques achieve high retrieval accuracy, their decision processes are largely opaque, as they operate directly on dense pixel-level features. Despite their successes, these methods remain tied to pixel-level appearance and degrade under strong viewpoint or environmental changes. Recent works attempt to mitigate this, such as PRGS \citep{Zuo2025PRGS}, which constructs patch-to-region graph representations to improve correspondence under appearance variation, and PairVPR \citep{Hausler2025PairVPR}, which introduces place-aware pre-training for transformer-based VPR.  
Parallel developments in vision–language modeling (e.g., CLIP \citep{Radford2021CLIP}, ALIGN \citep{Jia2021ALIGN}) have highlighted the power of textual representations for capturing semantic structure that remains stable despite appearance changes. From an intelligent systems perspective, language serves as an intermediate symbolic representation that bridges perception and reasoning. This has motivated text-driven localization frameworks such as Text2Pos \citep{Kolmet2022Text2Pos}, Text2Loc \citep{Xia2024Text2Loc}, and language-based scene retrieval \citep{Chen2024WhereAmI}. More recently, cross-modal approaches—including OPAL \citep{Kang2025OPAL}, Des4Pos \citep{Shang2025TextLidar}, and Text4VPR \citep{Shang2025TextVision}—demonstrate that textual or language-aligned embeddings can effectively support large-scale place recognition, leveraging the generalization ability of vision–language models and zero-shot retrieval \citep{Radford2021CLIP,Li2022BLIP}. Nevertheless, most existing language-based methods rely on unstructured embeddings, limiting their ability to explicitly reason about entities, relations, and scene structure.
Building upon these advances, we introduce Text2Graph VPR, an explainable semantic reasoning framework for place recognition that operates entirely in the semantic–graph domain. Given a sequence of images, our approach first generates detailed textual descriptions using a vision–language model (e.g., GPT-4V). These descriptions are parsed into scene graphs, in which nodes represent objects and attributes, and edges encode spatial or functional relations. This structured representation enables explicit reasoning over scene entities and their relationships, forming a compact and human-interpretable model of place semantics. Sequence-level graph aggregation captures both semantic consistency and structural context, yielding an appearance-invariant representation. To compare places, we propose a hybrid reasoning mechanism that combines:
\begin{itemize}
\item Learned graph embeddings obtained via a Graph Attention Network (GAT) \citep{Velickovic2018GAT}, which models semantic importance and relational dependencies between entities
\item Structural similarity computed using the Shortest-Path (SP) kernel \citep{Borgwardt2005SPKernel}, which explicitly measures topological alignment between scene graphs.
\end{itemize}
By fusing learned semantic relevance with explicit structural reasoning through an adaptive weighting strategy, the proposed system balances data-driven learning and symbolic comparison. Although Text2Graph VPR yields lower raw accuracy than some pixel-based baselines, it offers several properties that are central to intelligent expert systems:
\begin{itemize}
\item Semantic interpretability – decisions are grounded in explicit entities and relations.
\item Robust reasoning under appearance change – high-level semantics remain stable across lighting, seasonal, and weather variations.
\item Modality-independent inference– the system operates directly on text or graph inputs, enabling zero-shot text-only localization without retraining.
\item Cross-domain reasoning capability– semantic and relational abstractions generalize across cities and environments
\item Computational efficiency– compact graph-based representations support scalable inference.
\end{itemize}
In summary, our main contributions are:
\begin{enumerate}
\item An explainable semantic expert system for place recognition based on text-derived scene graphs.
\item A dual-similarity reasoning framework combining GAT-based semantic modeling with SP-kernel structural matching
\item A transparent and modality-flexible localization system supporting zero-shot textual reasoning.
\item Comprehensive experiments on Oxford RobotCar \citep{Maddern2017RobotCar} and MSLS \citep{Warburg2020MSLS} datasets, demonstrating competitive performance and strong generalization under cross-condition and cross-city settings.
\end{enumerate}
This work positions visual place recognition as a semantic reasoning problem rather than purely a perceptual matching task, bridging computer vision, natural language understanding, and graph-based inference. The remainder of the paper is organized as follows. Section 2 reviews related work on image- and sequence-based place recognition, language-driven localization, graph-based scene representations, and cross-modal retrieval. Section 3 describes the proposed methodology. Section 4 presents the experimental evaluation. Section 5 discusses limitations and future directions, and Section 6 concludes the paper.
\section{Related Work}
\label{title_page}
\subsection{Image-Based and Sequence-Based Visual Place Recognition}
Early visual place recognition (VPR) methods predominantly relied on handcrafted features such as SIFT or SURF. The advent of deep learning brought about a paradigm shift, with approaches like NetVLAD \citep{Arandjelovic2016NetVLAD} setting a strong foundation by aggregating local convolutional descriptors into global image embeddings. Building on this, methods such as PatchNetVLAD \citep{Hausler2021PatchNetVLAD} refined the approach by fusing both local and global information, while sequence-based systems like SeqNet \citep{Garg2021SeqNet} and DeepSeqSLAM \citep{Chancan2020DeepSeqSLAM} exploited temporal continuity to mitigate the challenges posed by significant appearance variations (e.g., due to weather, lighting, or seasonal changes). These methods introduced recurrent layers or temporal modeling to capture sequence continuity. More recently, JIST \citep{Berton2023JIST} and JVPR \citep{Li2023JVPR}, jointly train image- and sequence-level encoders, demonstrating that temporal aggregation acts as an implicit reasoning mechanism over sequential observations. These methods demonstrate that combining local frame-wise information with sequence context improves robustness under appearance change. More recent methods, including JVPR \citep{Li2023JVPR}, MR-NetVLAD \citep{Khaliq2022MultiResNetVLAD}, and TCGAT \citep{Wang2024MultimodalVPR}, further push the boundaries by incorporating multi-resolution strategies and, in some cases, additional modalities (such as depth or semantic segmentation) to enhance robustness. Sequence-based approaches, such as SeqMatchNet \citep{Garg2022SeqMatchNet} and SeqVLAD\citep{Mereu2022SequentialDescriptors} further highlight the benefit of spatio-temporal consistency as an inductive bias for decision-making in dynamic environments— particularly on urban datasets such as MSLS \citep{Warburg2020MSLS} and Oxford RobotCar \citep{Maddern2017RobotCar}. Despite these advances, most image- and sequence-based VPR methods rely on opaque representations learned directly from pixel data, limiting interpretability and controllability of their decisions. Recent work such as PRGS \citep{Zuo2025PRGS} introduces patch-to-region graph representations to improve correspondence under appearance changes, and PairVPR \citep{Hausler2025PairVPR} employs place-aware pre-training for transformer-based architectures.
\subsection{Text and Language for Localization}
Textual information has been effectively leveraged in various areas of computer vision and robotics to provide high-level semantic context. In tasks such as Visual Question Answering (VQA) and image captioning, natural language enables explicit reasoning about objects, attributes, and relationships beyond low-level visual cues. For instance, bottom-up and top-down attention mechanisms introduced by Anderson et al. \citep{Anderson2018BottomUp} and the Visual Genome dataset by Krishna et al. \citep{Krishna2017VisualGenome} introduced structured semantic supervision that bridges vision and language.
More recently, large-scale vision-language models like CLIP \citep{Radford2021CLIP} and ALIGN \citep{Jia2021ALIGN} further demonstrate that language supervision enables the learning of transferable and semantically meaningful visual representations. From an intelligent systems perspective, language serves as a symbolic abstraction layer that decouples perception from high-level reasoning, motivating its use in retrieval and localization tasks under severe appearance variation.
In localization, approaches such as Text2Pos \citep{Kolmet2022Text2Pos} and Text2Loc \citep{Xia2024Text2Loc} leverage natural language descriptions to localize agents within 3D environments, demonstrating that textual cues can guide spatial inference even when visual inputs change dramatically. MSSPlace \citep{Melekhin2025MSSPlace} and TextPlace \citep{Hong2019TextPlace} further integrate textual semantics in multi-sensor localization. The work “Where am I? Scene Retrieval with Language” \citep{Chen2024WhereAmI} shows that large language models can support spatial reasoning from free-form descriptions, enabling human-centric interaction with localization systems.
Recent works from 2023 to 2025—including LIGHTRAG \citep{Guo2024LightRAG}, LEXIS \citep{Kassab2024LEXIS}, Des4Pos \citep{Shang2025TextLidar}, and Text4VPR \citep{Shang2025TextVision}—reinforce the role of language as either a complementary or primary modality for spatial understanding. These studies collectively highlight that text provides robustness, interpretability, and cross-domain generalization—key properties for intelligent expert systems operating in real-world environments.
\subsection{Graph-Based Scene Representations}
Scene graphs have emerged as a powerful tool for modeling the complex relationships between objects within a scene. Traditionally used in tasks such as image captioning and scene understanding, graph-based representations are increasingly recognized as effective knowledge structures for reasoning-based perception systems.
Several studies have explored graph-based methods in localization and scene understanding, demonstrating the potential of graph neural networks to capture intricate inter-object relationships. For instance, Wan et al. \citep{Wan2022SceneGraphCompletion} proposed Representation Learning via Jointly Structural and Visual Embedding (RLSV), which jointly learns visual and structural embeddings for scene graph completion, showing improved performance in relational inference tasks. This integration of visual perception with structured reasoning is particularly relevant to spatial AI problems such as localization. 
In outdoor localization, GOTPR \citep{Jung2025GOTPR} constructs scene graphs from OpenStreetMap data and employs graph transformers for city-scale text-based geolocalization. Zha and Yilmaz \citep{Zha2023Subgraph} further extended the use of scene graphs by modeling motion trajectories as path subgraphs for topological geolocalization. Their approach encodes spatial movement through subgraph learning with GNNs, contributing a novel graph-based encoding for localization purposes. SceneGraphLoc \citep{Miao2024SceneGraphLoc}, proposed by Miao et al., explores cross-modal localization using 3D scene graphs that integrate geometric, semantic, and visual features. This method highlights how graph-based approaches can effectively handle multi-modal data and improve coarse localization. Complementing these, OPAL \citep{Kang2025OPAL} introduces visibility-aware graph-like fusion between LiDAR and OpenStreetMap structures, offering a multimodal perspective relevant to graph-enhanced spatial reasoning. However, most existing graph-based localization methods rely on geometric or map-derived graphs and rarely exploit free-form textual scene descriptions as a source of relational knowledge. This limits their ability to reason over semantic abstractions aligned with human language and explanations. 
\subsection{Zero-Shot Cross-Modal Retrieval and Semantic Localization}
The field of Visual Place Recognition (VPR) is increasingly exploring zero-shot cross-modal retrieval—a paradigm where systems retrieve visual or spatial information based on unseen textual queries, without requiring supervised training on that specific task. This approach leverages the semantic generalization capabilities of vision-language models, enabling localization even in the absence of direct visual inputs.
By embedding visual and textual modalities into a shared semantic space, models such as CLIP \citep{Radford2021CLIP} and ALIGN \citep{Jia2021ALIGN} enable generalization without task-specific supervision. Building upon this foundation, recent localization-specific works have begun exploring text-only queries for spatial retrieval. For example, Text2Pos \citep{Kolmet2022Text2Pos} and Text2Loc \citep{Xia2024Text2Loc} perform localization in 3D point clouds from natural language, without requiring paired image-text training. The “Where am I?” framework \citep{Chen2024WhereAmI} localizes users based on free-form scene descriptions and retrieves matching visual places using large language models and structured semantic representations. These approaches show that textual scene understanding can be grounded in space, offering robustness to appearance variations like lighting, weather, and seasonal shifts. Recent 2025 methods, such as Des4Pos \citep{Shang2025TextLidar} and Text4VPR \citep{Shang2025TextVision}, further strengthen this direction by demonstrating cross-modal retrieval in LiDAR and multi-view image settings, respectively, using language as a primary querying modality.
Zero-shot semantic localization is particularly attractive for intelligent expert systems, as it enables flexible querying, modality-independent reasoning, and human-interpretable interaction. These properties are crucial in safety-critical or visually degraded scenarios where raw imagery may be unreliable. A recent multimodal VPR survey by Li et al. \citep{Li2025MultimodalReview} provides a comprehensive overview of these trends and motivates further exploration of structured, language-grounded reasoning for spatial intelligence.
\begin{figure}[t!]
\centering
\includegraphics[width=\textwidth]{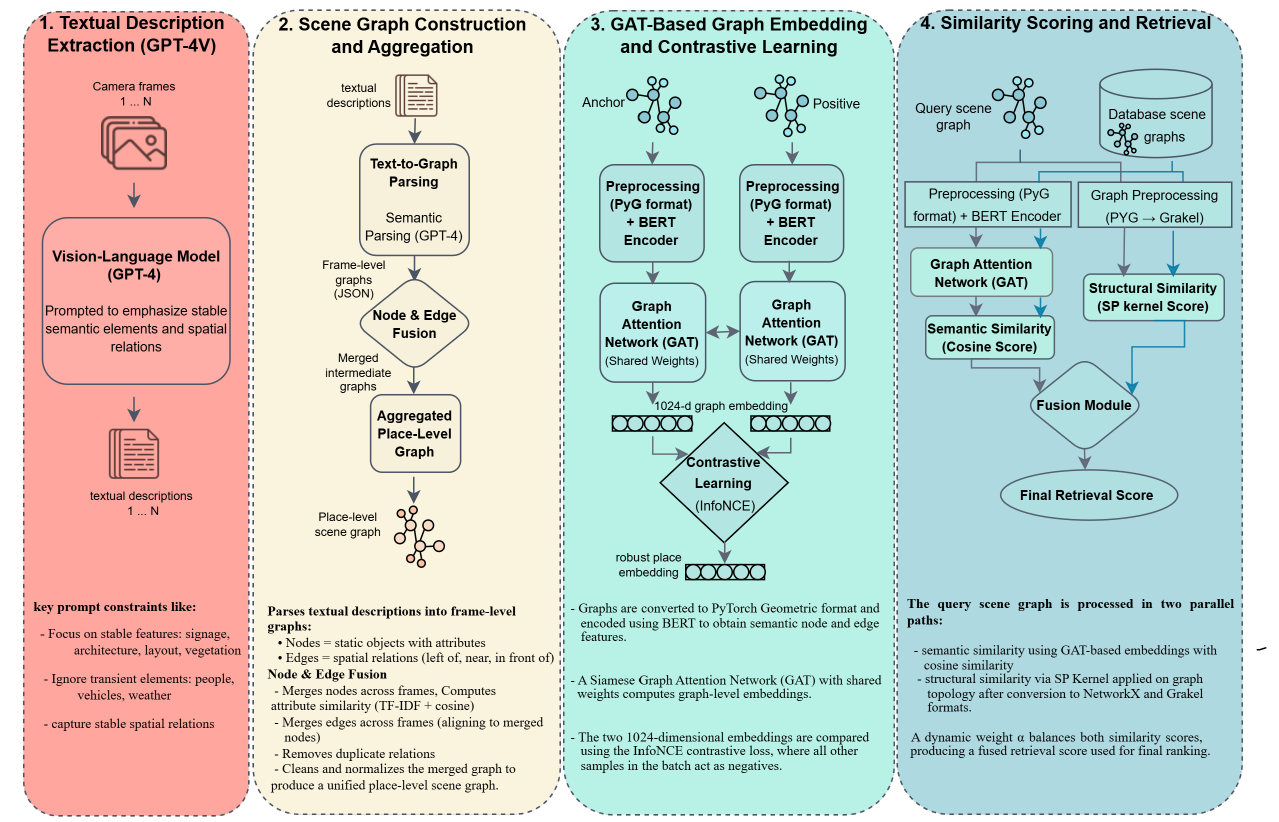}
\caption{Overview of the proposed text-to-graph Visual Place Recognition framework. The pipeline consists of four main stages. From left to right (1) A vision–language model generates stable textual descriptions from sequential frames. (2) A large language model parses these descriptions into frame-level scene graphs, which are subsequently merged into a unified place-level graph. (3) The scene graphs are encoded into graph embeddings using a Graph Attention Network (GAT) with BERT-based node and edge features, trained using an InfoNCE contrastive loss on anchor–positive pairs. (4) During retrieval, the query graph is compared with database graphs using a dual-similarity strategy: semantic similarity via cosine distance between learned embeddings, and structural similarity via the Shortest-Path (SP) kernel. A dynamic fusion weight balances the two signals to produce the final retrieval score.}\label{fig1}
\end{figure}
\section{Methodology}
We propose a novel text-to-graph-based framework for sequence-based Visual Place Recognition (VPR) that leverages vision-language models and structured graph representations to overcome challenges posed by appearance variations. In contrast to traditional VPR pipelines, our method is framed as an intelligent semantic reasoning system that transforms raw images into symbolic knowledge structures (textual scene graphs) and performs inference using both learned and rule-like structural cues. These graphs are then used for robust retrieval via a dual-similarity framework that combines learned embeddings and structural matching. Figure 1 illustrates the model of our proposed framework.
\subsection{Textual Description Extraction}
The first stage of our pipeline involves transforming each frame within an image sequence into a detailed natural language description using GPT-4V, a state-of-the-art vision-language model. To guide the model toward generating consistent and informative descriptions, we employ a carefully designed prompt that emphasizes semantically stable scene elements critical for place recognition. These include visible text or signage (e.g., street names or business names), architectural characteristics, urban layout (e.g., intersections or straight roads), infrastructure elements (e.g., traffic lights), and natural features such as trees and vegetation.
The prompt explicitly instructs the model to disregard transient or variable elements such as people, vehicles, lighting conditions, or weather-related phenomena, ensuring that the resulting descriptions remain invariant to environmental changes. Furthermore, it encourages spatial reasoning by requesting relative positioning of objects (e.g., “a tall brick building to the left of a narrow white house”), enabling a more structured interpretation of the scene. This stage functions as a knowledge-extraction module within an expert system, producing symbolic descriptions suitable for downstream inference. Figure 2a illustrates an example input image used in our pipeline, and Figure 2b shows the corresponding textual description generated in the first stage of our method.
\begin{figure}[t!]
\begin{center}
\centering
\includegraphics[width=\textwidth]{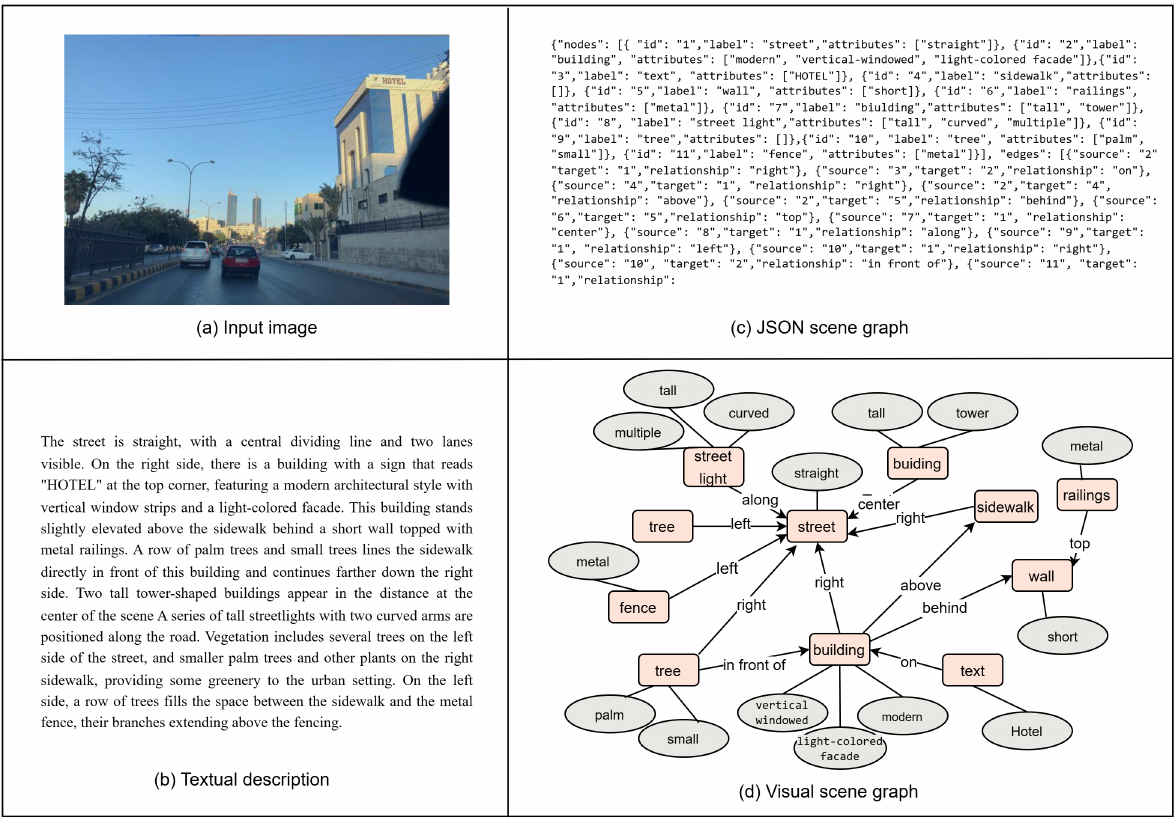}
\caption{Example of the scene graph construction pipeline. (a) Input image from an urban environment. (b) Generated textual description containing object-centric and spatial details. (c) JSON-formatted scene graph produced by the prompt-guided parsing process. (d) Visualization of the resulting scene graph showing object nodes and spatial relationships.}\label{fig2}
\end{center}
\end{figure}
\subsection{Scene Graph Construction and Aggregation}
To convert textual descriptions into structured semantic representations, we generate scene graphs where nodes denote static objects in the environment and edges represent spatial relationships between them. By converting language into structured symbolic graphs, this stage enables rule-like reasoning and supports explicit knowledge representation. This intermediate representation allows for consistent abstraction across varying visual appearances and supports graph-based reasoning downstream. To construct these graphs, we utilize a prompt-driven approach with a large language model (GPT-4) to parse natural language descriptions into structured JSON-formatted graphs. An example of the generated JSON scene graph structure is shown in Figure 2c. The prompt instructs the model to extract object-centric nodes (e.g., building, text, tree, traffic light) and attribute them with non-spatial properties (e.g., classical, light-colored, arched windows). Spatial relationships such as left, right, close by, or in front of are used to establish directed edges between object pairs. 
The prompt is designed to enforce consistent structure and vocabulary, ensuring that each scene graph adheres to a common schema and omits irrelevant or transient features (e.g., cars, people, or seasonal indicators). The corresponding graphical visualization of the scene graph is provided in Figure 2d.
Since each location in our dataset is represented by a sequence of frames, we construct one scene graph per frame and then aggregate them into a unified place-level representation. This is necessary to incorporate complementary semantic cues visible across different viewpoints within the same location.
To merge the individual scene graphs into a single, coherent representation, we implement a two-stage fusion strategy:
\begin{itemize}
\item Node merging: Nodes across frames are compared based on their object label and the semantic similarity of their attributes. We compute pairwise attribute similarity using TF-IDF vectorization and cosine similarity. If the similarity exceeds a predefined threshold (0.7), the nodes are merged, and their attributes are unioned.
\item Edge merging: Edges are similarly aggregated by aligning source and target nodes to the merged node identities and preserving unique spatial relationships. Duplicate edges are removed to ensure graph sparsity and interpretability.
\end{itemize}
This fusion process yields a robust and compact scene graph per place that integrates redundant object mentions and resolves minor inconsistencies in object descriptions across views. The resulting merged graph captures both the semantic content and spatial layout of the location, forming the input for graph embedding via GAT in the next stage.
\subsection{Graph Embedding with GAT}
To generate discriminative and robust embeddings from textual scene graphs, we employ a Graph Attention Network (GAT) architecture trained using a contrastive learning. Our approach consists of several stages: preparing the scene graph data for graph neural networks, encoding semantic features using pre-trained language models, and training a GAT model with a contrastive loss function to produce place-discriminative graph-level representations.
\subsubsection{Scene Graph Preprocessing and Feature Encoding}
Each scene graph is first preprocessed to extract its structure and semantic content. We begin by converting each JSON-formatted scene graph into a structured graph suitable for neural processing using PyTorch Geometric. Each graph contains a list of nodes and edges. For each node, we extract Label (e.g., "building", "tree") and Attributes (e.g., "tall"). In order to encode both identity and attribute-level information, we concatenate the object label with its associated attributes into a single phrase (e.g., “red metal mailbox”). This provides a more descriptive representation that preserves
both categorical and contextual details for each object in the scene. Edges represent relationships between object pairs—typically spatial or functional—such as “next to,” “on top of,” or “beside.” Each edge connects a source node and a target node, identified using unique node IDs. These relationship types are used as the edge features in the graph, and they serve to preserve the structural semantics of the scene layout.
Once the node and edge phrases are extracted, they are encoded using a pre-trained BERT-base model. This model transforms each textual phrase into a dense vector embedding that captures its semantic meaning. We apply mean pooling over the output hidden states from BERT to derive fixed-size embeddings for all nodes and edges. This ensures that the graph carries both semantic richness and consistency across varied scene descriptions. The final step in preprocessing involves converting the graph into a format compatible with PyTorch Geometric. This standardized format allows us to efficiently process graphs in batches during training and ensures compatibility with graph neural networks used in later stages. This pipeline enables each scene graph to act as a semantically grounded and structurally meaningful representation of a location, ready for downstream embedding and retrieval using our GAT-based model. This semantic encoding process transforms symbolic knowledge (nodes/edges) into dense reasoning vectors, enabling hybrid symbolic–neural inference consistent with intelligent expert systems.
\subsubsection{GAT Architecture and Attention Pooling}
To compute semantic embeddings from scene graphs, we employ a Graph Attention Network (GAT) \citep{Velickovic2018GAT} with explicit edge features. Each node representation is updated by attending to its neighbors through multi-head attention, where both node and edge attributes influence the message-passing process. Our encoder architecture is generally shallow to mitigate over-smoothing—a well-documented phenomenon in graph neural networks \citep{Rusch2023OversmoothingSurvey, Chen2020OverSmoothing}, where repeated message passing causes node embeddings to converge toward similar values, reducing discriminability. This effect is particularly severe in small and homogeneous scene graphs, which are typical in many urban sequences. In such graphs, feature collapse can occur after only a few propagation steps. We primarily adopt a two-layer GAT configuration, which provides sufficient contextualization while preserving node-level distinctiveness. However, as detailed in Section 4.3.4.3, the optimal depth varies across datasets depending on graph structure and scene diversity: smaller or more uniform graphs (e.g., Oxford, Amman) benefit from deeper aggregation to capture broader context, whereas larger and more heterogeneous graphs (e.g., San Francisco) perform best with a shallower configuration to avoid feature mixing.
To obtain a fixed-size representation for each scene graph, we apply global mean pooling, which aggregates node-level embeddings into a compact graph-level vector. The model is trained under an in-batch contrastive learning framework. Each batch contains paired scene graphs (anchor, positive) corresponding to the same place but observed under different viewpoints or conditions. In this setup, all other samples in the batch implicitly act as negatives, avoiding the need for explicit negative mining and leading to efficient utilization of training data. The objective follows the InfoNCE loss \citep{Oord2018CPC}, widely used in self-supervised representation learning, which encourages embeddings of anchor–positive pairs to be close in the learned space while simultaneously pushing apart embeddings of non-matching pairs. Formally, for a batch of size B, we first obtain $l_2$-normalized embeddings \begin{math}\left\{z_{i} \right\}^{2B}_{i=1} \end{math}, where each anchor–positive pair contributes two embeddings. The temperature-scaled cosine similarity between two embeddings $z_i$ and $z_j$ is defined as:
\begin{equation}
S_{ij} = \frac{\left\langle z_{i} , z_{j} \right\rangle}{\tau} \label{eq1}
\end{equation}
where $\tau$ $> 0$ is a temperature hyperparameter that controls the sharpness of the similarity distribution. For each anchor i, with its corresponding positive index $p(i)$, the InfoNCE contrastive loss is given by:
\begin{equation}
\mathcal{L}_{i} = -log\frac{exp(s_{i,p(i)}))}{\sum_{j\neq i}^{}exp(s_{ij})} \label{eq2}
\end{equation}
The final loss is averaged across all $2B$ samples in the batch:
\begin{equation}
\mathcal{L} = \frac{1}{2B} \sum_{i=1}^{2B}\mathcal{L}_{i}  \label{eq3}
\end{equation}
This in-batch formulation offers several advantages. By treating all non-matching samples in the batch as negatives, it naturally provides a large pool of negatives without explicit hard-negative mining, leading to stronger discriminative embeddings \citep{Chen2020SimCLR}. Moreover, InfoNCE has been shown to be theoretically connected to maximizing a lower bound on mutual information between positive pairs \citep{Poole2019MutualInfo}, ensuring that the model learns invariant and semantically meaningful features. Finally, the simultaneous attraction of anchor–positive pairs and repulsion of numerous negatives ensures that the learned embeddings exhibit invariance to viewpoint changes, robustness to environmental condition variations, and resistance to dataset-specific biases \citep{Oord2018CPC, Chen2020SimCLR, Khosla2020SupCon}.
Optimization was carried out using the Adam optimizer with a weight decay of $1*10^{-5}$ and an initial learning rate of $1*10^{-4}$, scheduled via cosine annealing for stable convergence. The network was trained for 500 epochs with a batch size of 128, ensuring a balance between gradient stability and GPU efficiency. The encoder produces a 1024-dimensional embedding vector that captures semantic entities and their relations, serving as the basis for downstream retrieval. All experiments were implemented in PyTorch \citep{Paszke2019PyTorch}.
\subsection{Structural Similarity via SP Kernel}
In addition to learning high-level semantic embeddings via graph neural networks, our framework incorporates a structural similarity measure to capture topological patterns within the scene graphs. Specifically, we employ the Shortest-Path (SP) kernel \citep{Borgwardt2005SPKernel}, a widely used graph kernel that computes similarity by analyzing the distribution of shortest paths between all pairs of nodes in two graphs. The SP kernel has several desirable properties for our application:
\begin{itemize}
    \item Invariance to node ordering: Since it relies on path lengths rather than node identities, it is unaffected by how nodes are indexed or labeled.
    \item Robustness to structural variation: It tolerates minor discrepancies in graph structure caused by occlusions, viewpoint changes, or incomplete object detection.
    \item Efficiency and positive definiteness: The kernel is computationally efficient and mathematically well-behaved, making it suitable for large-scale retrieval tasks and integration with kernel-based learning methods.
\end{itemize}

By comparing the topological layout of scenes—independently of specific node or edge features—the SP kernel complements our GAT-based embeddings, which focus primarily on semantic content. This dual approach allows our system to account for both what is in the scene and how it is arranged, improving robustness in scenarios where semantic or spatial cues alone may be insufficient.
In practice, for each query graph and candidate graph in the database, we compute the normalized SP kernel similarity and fuse it with the learned embedding similarity via a weighted combination (as detailed in Section 3.5). This enhances the system’s ability to retrieve correct matches even under challenging conditions such as partial observations or large appearance shifts. In the context of intelligent systems, the SP kernel acts as a deterministic structural reasoning component, providing interpretable rule-like cues about scene topology that complement learned semantic embeddings.
\subsection{Similarity Scoring and Retrieval}
Our retrieval strategy leverages a dual-similarity framework that integrates both semantic and structural information extracted from scene graphs. Rather than relying solely on visual features, we combine learned graph embeddings with topological analysis to achieve more robust place recognition under appearance and structural variations. This dual strategy allows our system to robustly match places even when one modality (semantic or structural) is insufficient alone, ensuring greater resilience to environmental changes, occlusions, and ambiguities in scene observation.
For semantic similarity, scene graphs are encoded into compact embeddings using the trained Graph Attention Network (GAT), which captures the contextual relationships between objects. Cosine similarity is computed between the query and database embeddings, measuring semantic alignment between scenes. For structural similarity, we analyze the underlying topology of scene graphs using the Shortest-Path (SP) kernel. The SP kernel compares the distribution of shortest paths within graphs, providing a measure that is invariant to node ordering and robust against small structural perturbations.
To adaptively balance these two complementary views, we introduce a dynamic weighting mechanism. The final retrieval score for a database entry is calculated as:
\begin{equation}
Similarity=\alpha*Semantic Similarity+(1-\alpha)*Structural Similarity \label{eq4}
\end{equation}
The choice of the weighting factor $\alpha$ is crucial, as it determines the balance between semantic (embedding-based) and structural (SP kernel-based) similarity. In our study, we investigated several strategies for selecting $\alpha$ and compared their performance. This dual-similarity fusion represents a hybrid reasoning strategy—combining learned semantic inference with symbolic structural reasoning—which is well aligned with intelligent decision-making systems.
\section{Experiments and Results}
\subsection{Datasets}
We evaluate our method on two widely adopted urban Visual Place Recognition (VPR) benchmarks: Oxford RobotCar \citep{Maddern2017RobotCar} and Mapillary Street-Level Sequences (MSLS) \citep{Warburg2020MSLS}. Both datasets encompass a wide range of challenging real-world conditions, including appearance variations caused by weather, illumination, seasonal changes, and viewpoint shifts, providing a realistic testbed for assessing the robustness of our semantic-based approach. For both datasets, frames are sampled such that the spatial distance between consecutive frames is approximately 2 meters, computed from geographic coordinates.\\
\textbf{Oxford RobotCar.} The Oxford RobotCar dataset is a large-scale collection of urban driving sequences captured across multiple traversals of the same route under diverse environmental conditions. For training, we use a single sequence recorded on an overcast day (2015-03-17-11-08-44). Evaluation is performed on four additional sequences, each exhibiting distinct variations in appearance (e.g., different lighting, weather, and time of day). Detailed information about the selected sequences for test is provided in Table 1.\\
\begin{table}[t]
\caption{Test sequences of Oxford dataset}
\centering 
\small
\begin{tabular}{|c|c c|}
\hline
     & Environment conditions	&Sequence selection  \\ \hline
Seq1 &Autumn day, Overcast       &2014-12-09-13-21-02 \\ \hline
Seq2 &Winter Night	             &2014-12-16-18-44-24 \\ \hline
Seq3 & Winter day, Snow	         &2015-02-03-08-45-10 \\ \hline
Seq4 &Summer day, Sun	         &2015-08-28-09-50-22 \\ \hline
\end{tabular}
\end{table}
\textbf{Mapillary Street-Level Sequences (MSLS).} The MSLS dataset consists of globally distributed street-level sequences collected from multiple cities, designed specifically for place recognition research. For training, we follow prior work such as SeqNet \citep{Garg2021SeqNet} and use a reduced subset of approximately 10,000 images from the city of Melbourne. For evaluation, we select sequences from Amman and San Francisco, which introduce distinct urban layouts and visual characteristics, enabling a thorough assessment of cross-city generalization. 
\subsection{Evaluation Protocol}
We evaluate our proposed method using the standard Recall@K (R@K) metric, commonly used in Visual Place Recognition (VPR). Recall@K measures the proportion of queries for which at least one of the top K retrieved database candidates lies within a specified localization radius from the ground-truth position. In our experiments, we use localization radii of 10 meters for the Oxford RobotCar dataset and 25 meters for the MSLS dataset, following standard practice. The choice of radius varies depending on the dataset characteristics and geographic scale. We report Recall@K across multiple values of K (e.g., K = 1, 5, 10, 20) to comprehensively evaluate retrieval performance under varying levels of difficulty.
\subsection{Designed Experiments and Results}
\subsubsection{Alpha Selection Strategies}
In formula (4), the weighting factor alpha ($\alpha$) is not predetermined but must be carefully selected, since different datasets and query graphs require different balances between semantic and structural similarity. In the context of an expert-system perspective, $\alpha$ acts as a decision-weighting parameter that balances two complementary sources of reasoning: semantic inference (GAT-based embedding similarity) and structural inference (SP-kernel topology matching). Thus, selecting $\alpha$ can be interpreted as a rule-selection problem within an intelligent decision-fusion module. To this end, we investigated three strategies for determining $\alpha$: \\
\textbf{a) Constant $\alpha$ (Fixed Weighting).} A straightforward baseline is to use a fixed value of $\alpha$ for all queries (e.g., 0.3, 0.5, 0.8). This approach ensures consistency and eliminates the need for query-dependent computation. As shown in Table 2, the choice of $\alpha$ substantially affects performance: on Oxford RobotCar, assigning a higher weight to semantic similarity ($\alpha$ = 0.8) produces the best results, indicating that semantic cues are dominant in this relatively homogeneous setting. Conversely, for the more diverse MSLS benchmark, lower semantic weights achieve superior performance (Table 3), suggesting that structural similarity contributes more effectively when graph variability and environmental heterogeneity are higher. To better interpret these differences, we analyzed the structural properties of scene graphs in both datasets—specifically, the number of nodes, average degree, and graph density—visualized in Figure 3. Oxford graphs tend to be smaller and denser, implying fewer detected objects but stronger inter-object connectivity, with a comparatively narrow distribution. This pattern reflects the architectural uniformity and consistent street layout of Oxford, where scene graphs across queries exhibit limited structural and semantic variation. Consequently, structural similarity offers limited additional discriminative power, and fixed weighting schemes favoring semantic cues are sufficient.\\
\begin{table*}[t]
\caption{Results on test sequences of the Oxford RobotCar dataset with different alpha selection strategies. Results are reported in Recall@k (\%).}
\label{tab:alpha_oxford}
\centering
\resizebox{\textwidth}{!}{%
\begin{tabular}{|l l| c c c c|}
\hline
\multicolumn{2}{|c|}{Alpha Selection Method} & Seq 1 & Seq 2 & Seq 3 & Seq 4 \\
\multicolumn{2}{|c|}{} & $1 / 5 / 20$ & $1 / 5 / 20$ & $1 / 5 / 20$ & $1 / 5 / 20$ \\
\hline
\multirow{3}{*}{Constant} 
& 0.3 & 32.4 / 61.3 / 74.5 & 10.2 / 43.8 / 50.3 & 31.7 / 61.7 / 73.4 & 29.2 / 58.4 / 65.7 \\
& 0.5 & 40.6 / 65.6 / 81.3 & 28.5 / \underline{63.3} / 75.1 & 35.1 / 71.6 / 82.3 & 31.2 / 70.8 / 79.5 \\
& 0.8 & \textbf{53.1} / \textbf{72.7} / \textbf{83.5} & \textbf{32.0} / \textbf{70.8}/ \textbf{82.3} & \textbf{43.5} / \textbf{78.8} / \textbf{90.2} & 31.5 / \textbf{74.0} / \textbf{87.4} \\
\hline
\multirow{2}{*}{Heuristic functions}
& Func1 (Threshold-Based) & \underline{46.5} / \underline{70.2} / \underline{82.8} & \underline{30.5} / 59.5 / \underline{76.4} & 33.4 / \underline{73.1} / \underline{84.1} & 30.8 / \underline{73.1} / \underline{87.0} \\
& Func2 (Linear Weighted Logistic) & 43.2 / 69.4 / 79.4 & 23.5 / 50.4 / 65.2 & \underline{41.6} / 67.3 / 87.2 & \underline{35.8} / 68.5 / 81.3 \\
\hline
\multirow{2}{*}{Learning-based models}
& Model1 (MLP) & 35.2 / 61.6 / 73.1 & 21.9 / 48.3 / 59.2 & 32.1 / 54.8 / 69.5 & 28.1 / 57.1 / 62.5 \\
& Model2 (XGBRegressor) & 37.7 / 63.5 / 75.2 & 21.1 / 50.4 / 62.2 & 33.1 / 65.3 / 76.4 & 29.7 / 58.5 / 61.4 \\
\hline
\end{tabular}
}
\vspace{1mm}
\noindent\textit{Note:} The best overall results are shown in bold, while the second-best results are underlined. Results with the best-performing GAT configuration from the ablation study (Oxford results are obtained using a 2-layer GAT encoder).  
\end{table*}
 \begin{table}[t]
\caption{Results on Amman and San Francisco cities of the MSLS dataset with different alpha selection strategies. Results are reported in Recall@k (\%).}
\label{tab:alpha_msls}
\centering
\resizebox{\textwidth}{!}{%
\small
\begin{tabular}{|l l| c c|}
\hline
\multicolumn{2}{|c|}{Alpha Selection Method} & Amman & San Francisco \\
\multicolumn{2}{|c|}{} & $1 / 5 / 20$ & $1 / 5 / 20$ \\
\hline
\multirow{3}{*}{Constant}
& 0.3 & \underline{22.6} / \textbf{51.4} / \textbf{83.8} & 20.5 / 45.6 / 59.4 \\
& 0.5 & 21.6 / 43.2 / 73.0 & \textbf{27.4} / \textbf{48.3} / \textbf{66.7} \\
& 0.8 & 18.3 / 35.9 / 70.3 & 17.8 / 42.9 / 57.2 \\
\hline
\multirow{2}{*}{Heuristic functions}
& Func1 (Threshold-Based) & 21.3 / \underline{45.0} / \underline{73.3} & \underline{21.5} / \underline{47.1} / \underline{64.5} \\
& Func2 (Linear Weighted Logistic) & \textbf{24.5} / 40.5 / 67.6 & 20.7 / 45.4 / 59.8 \\
\hline
\multirow{2}{*}{Learning-based models}
& Model1 (MLP) & 18.4 / 43.2 / 75.7 & 17.0 / 40.4 / 54.2 \\
& Model2 (XGBRegressor) & 21.5 / 40.5 / 67.6 & 17.5 / 43.5 / 59.4 \\
\hline
\end{tabular}
}
\vspace{1mm}
\noindent\textit{Note:} The best overall results are shown in bold, while the second-best results are underlined. Results with the best-performing GAT configuration from the ablation study (Amman: 2 layers; San Francisco: 1 layer).  
The best and second-best results for each dataset are highlighted. 
\end{table}
 In contrast, MSLS graphs are larger, sparser, and more widely distributed across node count, degree, and density. This variability arises from MSLS’s greater geographic spread and semantic diversity. In such conditions, appearance-based semantic matching alone becomes inadequate, and structural similarity provides complementary information. As a result, smaller values of $\alpha$ (e.g., 0.3–0.5), which place greater emphasis on structural similarity, yield improved retrieval performance.\\
\begin{figure}[t]
\begin{center}
\centering
\includegraphics[width=\textwidth]{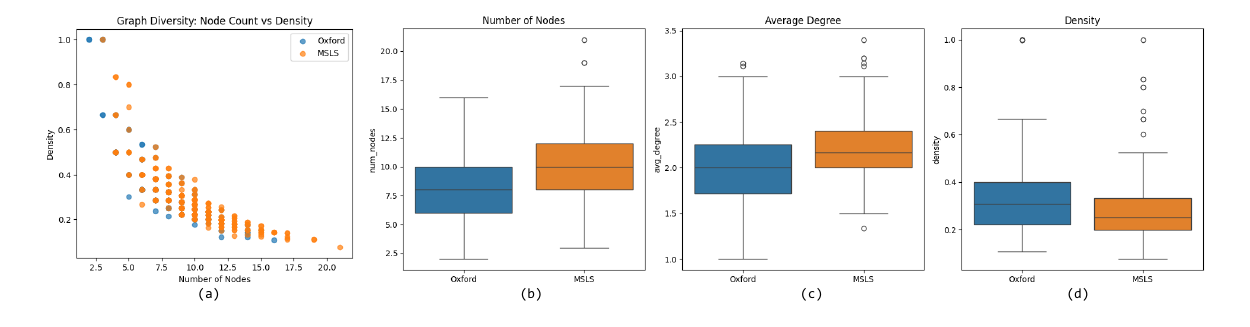}
\caption{Structural analysis of scene graphs in Oxford RobotCar and MSLS. From left to right: (a) Node Count vs. Density shows smaller, denser graphs in Oxford and larger, sparser graphs in MSLS. (b) MSLS graphs contain more nodes. (c) MSLS graphs have slightly higher average degree. (d) Density distribution further confirms Oxford’s higher compactness and MSLS’s sparsity. The plots show that Oxford graphs are smaller and denser. In contrast, MSLS graphs are larger but sparser.}\label{fig3}
\label{fig:pipeline}
\end{center}
\end{figure}
\textbf{(b) Feature-Based Adaptive Functions.} To move beyond fixed weighting, we investigated adaptive fusion strategies that map query-graph properties to an $\alpha$ value. The rationale was that small or sparse graphs (few nodes, low degree, low density) tend to carry weaker structural evidence and should therefore rely more on semantic embeddings (higher $\alpha$), whereas larger or denser graphs are more structurally informative and should lean more on kernel similarity (lower $\alpha$). Following this intuition, we designed two lightweight heuristic functions that take simple features of each query graph and map them to $\alpha$. The first approach, Threshold-Based Rules, applies a series of hand-crafted conditional thresholds on node count, average degree, and density. This rule-based function is intuitive, computationally inexpensive, and fully interpretable, as each decision reflects explicit structural cues in the graph. The second approach, Linear Weighted Logistic, defines $\alpha$ as a logistic function of the number of nodes and average degree, using a linear combination followed by a sigmoid activation. This formulation allows smooth, continuous adjustment of $\alpha$ across different graph configurations rather than abrupt threshold transitions. As reported in Table 2 and Table 3, the Threshold-Based Rule achieved consistently better results than the logistic variant across both datasets and its estimated $\alpha$ values closely matched the best-performing constant $\alpha$.\\
Adaptive fusion is particularly effective when fused modalities provide complementary, condition-dependent information, and when there are reliable per-query signals to weigh their reliability. Prior work in VPR and robotics has demonstrated this clearly: AdaFusion adaptively fuses visual and LiDAR features via learned attention \citep{Lai2022AdaFusion}, and MinkLoc++, which employs a late fusion of image and LiDAR descriptors for place recognition \citep{Komorowski2021MinkLoc}. The broader concept of uncertainty-based weighting—originally applied in multi-task learning via homoscedastic uncertainty estimation—also supports the intuition of balancing modality contributions based on confidence \citep{Kendall2018MultiTask}. These studies highlight that adaptive fusion is powerful when modality reliability fluctuates across queries.\\ 
Here, fusion refers to combining semantic and structural similarities, and adaptive fusion indicates dynamically adjusting the relative weighting ($\alpha$) for each query. In our case, however, both semantic embedding and structural similarity are derived from the same scene graph. While they emphasize different aspects—node semantics vs. graph topology—they remain correlated since they arise from the same graph structure. This correlation limits the utility of adaptive $\alpha$ because adjustments in $\alpha$ end up reinforcing highly overlapping information.\\
\textbf{(c) Learning-Based Prediction.} We also explored learning-based approaches to directly predict the weighting parameter $\alpha$ from query-graph features. Specifically, we extracted graph-level descriptors including node count, average degree, density, average shortest path length, and the mean embedding norm. These features served as inputs (x), while the best $\alpha$ values obtained from grid search were used as targets (y). Several regression and classification models were evaluated, including Gradient Boosting, Random Forest, XGBoost, and MLP architectures. The results for the two top-performing regressors—MLP and XGBRegressor—are reported in Tables 2 and 3 for the Oxford and MSLS datasets.\\
In practice, regression models produced more consistent and accurate $\alpha$ predictions than classification models. Since $\alpha$ is inherently a continuous variable, regression preserved its fine-grained variability across queries, whereas classification discretized $\alpha$ into coarse bins, reducing sensitivity and introducing quantization errors. Empirical results 
(Tables 2 and 3) confirm that learning-based prediction lagged behind both fixed and adaptive heuristics. Two key limitations contributed to this outcome: (i) the restricted feature space—comprising only global graph statistics—did not sufficiently encode the complex semantic–structural relationships influencing the optimal $\alpha$; and (ii) the limited amount of training data hindered generalization, causing regressors to partially collapse toward average $\alpha$ values (typically around 0.3–0.5). Overall, although regression offers a principled and potentially extensible solution, in our setting learning-based $\alpha$ prediction proved unreliable and less robust than non-learned adaptive approaches.
\subsubsection{Comparison with State-of-the-Art Methods}
Unlike purely visual systems, our framework performs a form of semantic-graph reasoning, interpreting objects and relations as knowledge units. This shifts the task from appearance matching toward symbolic inference, aligning with the principles of intelligent decision-making systems. The results in Tables 4 and 5 compare our method against several state-of-the-art baselines on the Oxford RobotCar and MSLS benchmarks.\\
\begin{table}[t]
\caption{Comparison of existing methods and the proposed Text2Graph VPR on the Oxford RobotCar dataset, reported in terms of Recall@k (\%).}
\centering
\resizebox{\textwidth}{!}{%
\small
\begin{tabular}{|l|cccc|}
\hline
Method & Seq1 & Seq2 & Seq3 & Seq4 \\
      & 1 / 5 / 20 & 1 / 5 / 20 & 1 / 5 / 20 & 1 / 5 / 20 \\
\hline
NetVLAD \citep{Arandjelovic2016NetVLAD}     & \textbf{65.7} / \textbf{76.3} / 82.3 & 46.8 / 69.6 / 97.4 & 64.0 / 76.9 / 83.2 & 83.3 / 92.5 / 96.3 \\
SeqNet \citep{Garg2021SeqNet}                & 65.8 / 72.6 / 83.2 & 74.1 / 87.5 / 97.6 & \textbf{65.7} / 73.2 / 82.8 & \textbf{93.0} / 96.0 / 98.2 \\
PatchNetVLAD \citep{Hausler2021PatchNetVLAD}& 58.3 / 72.9 / 81.0 & 73.9 / 89.0 / 96.5 & 59.2 / 75.2 / 82.5 & 71.8 / 88.2 / 95.3 \\
AP-GeM \citep{Revaud2019APLoss}               & 60.5 / 70.7 / 79.6 & 28.6 / 49.8 / 70.9 & 56.3 / 69.5 / 79.9 & 62.1 / 77.0 / 86.8 \\
MR-NetVLAD \citep{Khaliq2022MultiResNetVLAD}           & \textbf{65.7} / 76.0 / \textbf{85.0} & 91.1 / 97.7 / 98.6 & 65.2 / 77.5 / 83.2 & 88.9 / 94.0 / \textbf{98.6} \\
TCGAT \citep{Wang2024MultimodalVPR}                     & 66.4 / 76.4 / 83.4 & \textbf{93.6} / \textbf{98.0} / \textbf{99.6} & 63.4 / 76.6 / 83.5 & \textbf{93.0} / \textbf{96.8} / 98.2 \\
\textbf{Ours (Text2Graph VPR)}              & 53.1 / 72.7 / 83.5 & 32.0 / 70.8 / 82.3 & 43.5 / \textbf{78.8} / \textbf{90.2} & 31.5 / 74.0 / 87.4 \\
\hline
\end{tabular}
}
\label{tab:oxford_comparison}
\end{table}

\textbf{Oxford RobotCar:} To evaluate the effectiveness of our text-to-graph–based retrieval framework, we benchmark against a range of leading image- and sequence-based methods. AP-GeM \citep{Revaud2019APLoss} employs generalized mean pooling (GeM) with a listwise ranking loss to directly optimize mean Average Precision (mAP). NetVLAD \citep{Arandjelovic2016NetVLAD} uses differentiable VLAD pooling to aggregate local features into compact global descriptors. SeqNet \citep{Garg2021SeqNet} enhances retrieval through temporal re-ranking, generating sequence-level representations by aggregating frame-level descriptors. PatchNetVLAD \citep{Hausler2021PatchNetVLAD} improves spatial precision by introducing patch-level feature matching, while MR-NetVLAD \citep{Khaliq2022MultiResNetVLAD}  incorporates multi-resolution representations using image pyramids to capture features at multiple scales. Finally, TCGAT \citep{Wang2024MultimodalVPR} integrates spatial and temporal reasoning via a time-constrained graph attention mechanism, enabling joint modeling of sequential descriptors.\\
Table 4 presents a detailed comparison across four challenging Oxford RobotCar test sets: overcast–night, overcast–overcast (day), overcast–summer (day), and overcast–winter (day). These settings include strong appearance and illumination variations representative of real-world urban environments. Our model achieves Recall@20 scores of 83.5\%, 82.3\%, 90.2\%, and 87.4\% across the four test conditions—comparable to, or surpassing, several established baselines. Although the proposed method does not consistently yield the highest Recall@1, it maintains robust performance across Recall@5 and Recall@20, indicating stable retrieval within a small candidate set. This consistency highlights the method’s potential suitability for downstream refinement or re-ranking stages in sequence-based localization pipelines. Although our model operates on abstract semantic–structural representations rather than raw pixel features, it generalizes effectively across diverse appearance conditions by jointly exploiting stable semantic and relational cues from sequential scene graphs.\\
\begin{table}[t]
\caption{Comparison of existing methods and the proposed Text2Graph VPR on the MSLS dataset, reported in terms of Recall@k (\%). A dash (--) indicates metrics not reported in the original paper.}
\centering
\resizebox{\textwidth}{!}{%
\small
\begin{tabular}{|l|cc|}
\hline
Method & Amman & San Francisco \\
       & 1 / 5 / 10 / 20 & 1 / 5 / 10 / 20 \\
\hline
NetVLAD~\citep{Arandjelovic2016NetVLAD}      & 18.9 / 25.1 / 27.7 / -- & 28.9 / 39.8 / 45.5 / -- \\
SeqMatch~\citep{Garg2022SeqMatchNet}               & 24.6 / 30.2 / 33.0 / -- & 36.3 / 43.0 / 46.0 / -- \\
SeqNet~\citep{Garg2021SeqNet}                & 26.9 / 37.6 / 40.8 / -- & 55.6 / 67.1 / 72.8 / -- \\
HVPR~\citep{Garg2021SeqNet}                       & 27.0 / 33.0 / 35.0 / -- & 43.0 / 51.0 / 55.0 / -- \\
SeqVLAD~\citep{Mereu2022SequentialDescriptors}                 & 30.0 / 44.8 / 51.9 / -- & 66.1 / 82.6 / 86.3 / -- \\
Spatio-Temporal-SeqVPR~\citep{Zhao2024STAVPR} & \textbf{30.3} / 42.3 / 51.1 / -- & \textbf{68.0} / \textbf{84.1} / 86.4 / -- \\
JVPR~\citep{Li2023JVPR}                       & 25.0 / 32.0 / 35.0 / -- & 39.0 / 50.0 / 55.0 / -- \\
Delta descriptors~\citep{Garg2020DeltaDescriptors}         & 24.0 / 30.0 / 32.0 / -- & 34.0 / 43.0 / 47.0 / -- \\
PatchNetVLAD~\citep{Hausler2021PatchNetVLAD}& 25.4 / 33.9 / -- / 39.6 & 46.3 / 62.0 / -- / 72.8 \\
AP-GeM~\citep{Revaud2019APLoss}               & 12.2 / 20.3 / -- / 33.4 & 31.7 / 45.8 / -- / 58.2 \\
MR-NetVLAD~\cite{Khaliq2022MultiResNetVLAD}           & 22.9 / 31.0 / -- / 36.6 & 48.8 / 62.6 / -- / 73.2 \\
TCGAT~\cite{Wang2024MultimodalVPR}                     & 28.7 / 34.5 / -- / 41.9 & 62.3 / 75.1 / -- / \textbf{82.9} \\
\textbf{Ours (Text2Graph VPR)}              & 22.6 / \textbf{51.4} / \textbf{59.5} / \textbf{83.8} & 27.4 / 48.3 / 57.6 / 66.7 \\
\hline
\end{tabular}
}
\label{tab:msls_comparison}
\end{table}

\textbf{Mapillary Street-Level Sequences (MSLS):} For MSLS, we extend our comparison to include a broader range of sequence-based approaches. Among traditional sequence-based techniques, we include HVPR \citep{Garg2021SeqNet} and SeqMatchNet \citep{Garg2022SeqMatchNet}, both of which rely on frame sequence matching. We also report results from Delta Descriptors \citep{Garg2020DeltaDescriptors}, a non-learned aggregation technique, and SeqVLAD \citep{Mereu2022SequentialDescriptors}, which provides a detailed taxonomy of sequence pooling strategies and explores Transformer-based sequence encoding. Additional baselines include JVPR \citep{Li2023JVPR} and Spatio-Temporal-SeqVPR \citep{Zhao2024STAVPR}: JVPR jointly trains global and sequential descriptors to enhance their mutual consistency and robustness, while Spatio-Temporal-SeqVPR employs spatio-temporal attention to generate highly discriminative sequence-level representations that capture both spatial structure and temporal dynamics. \\
In line with prior work, we report Recall@1, @5, and either @10 or @20 depending on the availability of published metrics. Since different papers report varying cutoff thresholds, all available results are included for completeness (see Table 5). Our Text2Graph VPR method achieves competitive and stable performance across both Amman and San Francisco splits. Despite operating with compact textual–graph descriptors and being trained only on the simplified Melbourne subset (a smaller, less diverse portion of the original MSLS dataset), our model demonstrates strong generalization and robustness under significant appearance and viewpoint variations. Notably, it attains Recall@5 = 51.4\% and Recall@20 = 83.8\% on Amman, which are comparable to or exceed several pixel-based baselines such as PatchNetVLAD \citep{Hausler2021PatchNetVLAD}, AP-GeM \citep{Revaud2019APLoss}, and MR-NetVLAD \citep{Khaliq2022MultiResNetVLAD}, despite their higher-dimensional visual embeddings and access to richer training data. These findings highlight that semantic–structural reasoning via scene graphs provides a compact yet discriminative representation capable of generalizing effectively across geographically and visually diverse environments—even when trained on limited data.\\
We observe a clear performance gap between the Oxford RobotCar and the more challenging MSLS splits (Amman and San Francisco). While previous methods often reach recall above 90\% on Oxford by exploiting appearance-rich visual cues within the same city, their accuracy drops markedly on MSLS, frequently below 60\%. In contrast, our Text2Graph VPR achieves comparable or superior performance on MSLS, despite relying on semantic scene representations rather than pixel-level appearance. This suggests that our graph-based abstraction contributes to stronger robustness under cross-city and cross-condition variations. Such robustness, achieved through semantic reasoning rather than pixel features, demonstrates that the system generalizes through knowledge abstraction — a desirable property for intelligent expert systems.
\begin{table}[t]
\caption{Comparison of model complexity and efficiency across representative sequence-based VPR methods.}
\centering
\resizebox{\textwidth}{!}{%
\small
\begin{tabular}{|l|cccccc|}
\hline
Method & Backbone & Descriptor Dim. & Params (M) & Extraction (ms/seq) & Matching (ms) & GPU Mem. (GB) \\
\hline
SeqNet~\citep{Garg2021SeqNet}                & VGG-16   & 4096 & 83.89  & 31.0 & 1.1 & 2.68 \\
NetVLAD~\citep{Arandjelovic2016NetVLAD}     & ResNet-18 & 4096 & 11.7   & 30.4 & 1.1 & 2.26 \\
SeqMatch~\citep{Garg2022SeqMatchNet}               & VGG-16   & 4096 & 14.74  & 39.8 & 500.1 & 2.68 \\
SeqVLAD~\citep{Mereu2022SequentialDescriptors}                 & ResNet-18 & 4096 & 7.15   & 17.2 & 1.1 & 2.19 \\
Spatio-Temporal-SeqVPR~\citep{Zhao2024STAVPR} & CCT384   & 4096 & 13.06  & 18.7 & -- & -- \\
Delta descriptors~\citep{Garg2020DeltaDescriptors}         & VGG-16   & 4096 & 14.74  & 39.8 & 1.1 & 2.68 \\
HVPR~\citep{Garg2021SeqNet}                       & VGG-16   & 4096 & 14.74  & 39.8 & 0.9 & 2.68 \\
Jist~\citep{Berton2023JIST}                  & ResNet-18 & 512  & 11.7   & 11.1 & \textbf{0.1} & 2.04 \\
\textbf{Ours (Text2Graph VPR)}             & GAT      & 1024 & $\sim$\textbf{2.4} & \textbf{2.1} & 0.4 & $\sim$\textbf{1} \\
\hline
\end{tabular}
}
\label{tab:complexity}
\end{table}

\subsubsection{Computational cost}
\textbf{Training time.} Our model is lightweight and can be trained efficiently on a single GPU. With only $\approx$ 2.4 M parameters, it is 6×-30× smaller than most sequence-based baselines such as NetVLAD (14.7 M) and SeqNet (83.9 M).
Thanks to the compact GAT-based encoder and low-dimensional textual scene-graph inputs, training converges rapidly—typically within $\approx$ 1-1.5 hours on a single NVIDIA T4 GPU (Google Colab). In contrast, CNN-based architectures like SeqNet \citep{Garg2021SeqNet} and NetVLAD \citep{Arandjelovic2016NetVLAD} require substantially longer training due to their high-resolution image inputs and heavy convolutional backbones (e.g., ResNet-50/101). The reduced model complexity makes our approach highly practical for rapid experimentation, cross-city fine-tuning, and resource-constrained deployment (see Table 6).\\
\textbf{Inference time.} In sequence-based VPR, inference consists of two stages: (i) descriptor extraction, where the query scene graph is encoded into an embedding, and (ii) matching, where the query embedding is compared against the database to retrieve the closest candidates. Descriptor extraction in our framework is highly efficient, as scene graphs are compact (typically 6–16 nodes) and the GAT encoder is lightweight. The average extraction time is $\approx$ 2.1 ms per 5-frame sequence (forward pass only), which is significantly faster than CNN-based methods relying on high-resolution imagery. For completeness, the end-to-end pipeline—including optional caption generation and scene-graph parsing—takes $\approx$ 8–10 ms per sequence, still within real-time range for mobile-robot perception.
Matching is also efficient due to the compact 1024-D embeddings, compared to 2048-D descriptors typically used in CNN-based VPR. While extraction time remains constant per query, matching time scales linearly with database size, making compact embeddings crucial for scalability. Overall, our approach provides faster inference and lower computational cost, particularly in large-scale retrieval scenarios (see Table 6).\\
\textbf{Memory footprint.} Efficient retrieval requires storing all database descriptors in memory. The overall memory demand is proportional to $N_{db}*D$, where $N_{db}$ is the number of stored graphs and D is the descriptor dimension. Because our embeddings are only 1024-D (compared with 2048–4096 D in image-based approaches), the memory footprint is 2×–4× smaller, even for large-scale databases. This reduction not only decreases storage requirements but also accelerates nearest-neighbor search, as smaller vectors yield faster distance computations. \\
In summary, the combination of fast training, lightweight inference, and compact embeddings makes the proposed Text2Graph VPR model computationally efficient and well-suited for large-scale, real-time place recognition in robotics and autonomous-navigation applications (see Table 6). The lightweight nature of the model also aligns with the design of practical expert systems, which traditionally emphasize efficient rule evaluation and knowledge-based inference without heavy computational overhead.
\subsubsection{Ablation Studies}
\textbf{Model Flexibility to Sequence Length (Conceptual Discussion).} While empirical evaluation under variable sequence lengths was not conducted due to API usage costs, we discuss this property conceptually. In real-world VPR, the number of available frames per query may vary depending on vehicle speed, camera rate, or environmental factors. Methods whose descriptor dimensionality scales with sequence length (e.g., concatenation-based approaches) require re-training when the input length changes. In contrast, the proposed Text2Graph VPR framework abstracts each frame into a textual description and merges them into a unified scene graph. This process yields a fixed-dimensional representation determined by semantic content rather than sequence size. Consequently, the model can theoretically process both shorter and longer sequences without architectural modifications. This property enhances generalization and practical deployability, which we plan to evaluate quantitatively in future work.\\
\begin{table}[t]
\caption{Robustness to frame reversal on the MSLS dataset.}
\centering
\resizebox{\textwidth}{!}{%
\small
\begin{tabular}{|l|cccc|}
\hline
Method & Backbone & Forward & Backward & Diff (\%) \\
\hline
SeqNet~\citep{Garg2021SeqNet}            & VGG-16   & 50.1 & 42.0 & $-16\%$ \\
NetVLAD~\citep{Arandjelovic2016NetVLAD} & ResNet-18 & 68.5 & 65.1 & $-5\%$ \\
SeqMatch~\citep{Garg2022SeqMatchNet}           & VGG-16   & 44.8 & 24.2 & $-46\%$ \\
Delta descriptors~\citep{Garg2020DeltaDescriptors}     & VGG-16   & 43.0 & 11.7 & $-73\%$ \\
HVPR~\citep{Garg2021SeqNet}                   & VGG-16   & 51.0 & 28.5 & $-44\%$ \\
SeqVLAD~\citep{Mereu2022SequentialDescriptors}             & ResNet-18 & 85.5 & 85.2 & $-0.4\%$ \\
SeqVLAD~\citep{Mereu2022SequentialDescriptors}             & CCT384   & 89.2 & 89.2 & $\textbf{0.0\%}$ \\
TimeSformer~\citep{Bertasius2021TimeAttention}     & --       & 81.5 & 81.5 & $\textbf{0.0\%}$ \\
Jist~\citep{Berton2023JIST}               & ResNet-18 & 90.6 & 90.6 & $\textbf{0.0\%}$ \\
\textbf{Ours (Text2Graph VPR)}          & GAT      & 25.4 & 25.4 & $\textbf{0.0\%}$ \\
\hline
\end{tabular}
}
\label{tab:frame_reversal}
\end{table}
\textbf{Effect of reversing frames.} Robustness to frame ordering is essential for real-world VPR, where the same route may be traversed in opposite directions. To evaluate this, we compare performance when the frame sequence is presented in forward versus reversed order (Table 7). Methods such as SeqVLAD \citep{Mereu2022SequentialDescriptors} and TimeSformer \citep{Bertasius2021TimeAttention}, which perform early or holistic temporal fusion, exhibit inherent order invariance and maintain stable performance under reversal. In contrast, approaches relying on sequential alignment or late temporal aggregation, such as SeqMatchNet \citep{Garg2022SeqMatchNet} or HVPR \citep{Garg2021SeqNet}, suffer substantial degradation when frame order is reversed.\\
Our proposed Text2Graph VPR is naturally invariant to frame order through its semantic–structural abstraction. Each frame is first converted into a textual description and parsed into a scene graph; these per-frame graphs are then merged into a unified representation based on object–relation consistency rather than temporal order. Consequently, the final place-level descriptor remains identical even if the sequence is reversed. This property ensures robust bidirectional localization without the need to store or process separate forward and backward database representations.
For comparison with prior studies that report results on the MSLS Copenhagen and San Francisco splits, we report our results on the Amman and San Francisco splits. As the reversal difference (Diff) is the primary metric of interest, we report the mean Recall@1 across the two cities for our method.\\
\begin{table}[t]
\caption{Effect of GAT depth on retrieval performance on the MSLS dataset and two sequences of the Oxford RobotCar dataset.}
\centering
\resizebox{\textwidth}{!}{%
\small
\begin{tabular}{|l|cccc|}
\hline
GAT Layers & Oxford(Seq1) & Oxford(Seq2) & Amman & San Francisco \\
          & 1 / 5 / 20 & 1 / 5 / 20 & 1 / 5 / 20 & 1 / 5 / 20 \\
\hline
N = 1 & 35.7 / 60.2 / 73.9 & 22.5 / 48.6 / 67.2 & 15.6 / 32.7 / 70.9 & \textbf{27.4 / 48.3 / 66.7} \\
N = 2 & \textbf{53.1 / 72.7 / 83.5} & \textbf{32.0 / 70.8 / 82.3} & \textbf{22.6 / 51.4 / 83.8} & 22.4 / 42.4 / 62.7 \\
N = 3 & 44.6 / 66.1 / 74.8 & 27.4 / 60.7 / 68.8 & 21.4 / 50.6 / 80.3 & 17.2 / 34.8 / 50.7 \\
\hline
\end{tabular}
}
\label{tab:gat_depth}
\end{table}

\textbf{Effect of GAT Depth (Over-smoothing Ablation).}
We conducted an ablation study to analyze the effect of GAT depth on scene graph representation learning. Table 8 summarizes the retrieval performance obtained with different depths configurations. The best overall performance was obtained with a two-layer GAT, while adding more layers substantially degraded accuracy. This degradation is attributed to the over-smoothing phenomenon \citep{Rusch2023OversmoothingSurvey, Chen2020OverSmoothing}, where repeated message passing causes node embeddings to converge toward similar values, diminishing discriminability. The effect is particularly pronounced in our setting because scene graphs are small and contain limited neighborhood diversity, making them prone to rapid feature collapse. \\
Across datasets, the optimal GAT depth varies with graph structure and scene diversity. In the Oxford RobotCar and MSLS–Amman subsets—where graphs are relatively small and semantically coherent—two layers provide the best Recall@1 by capturing second-order context without inducing over-smoothing. In contrast, for the larger and more heterogeneous San Francisco subset, the one-layer model performs best. Here, additional propagation does not cause classical over-smoothing but rather feature over-mixing, where messages from semantically diverse or weakly related regions introduce noise that blurs contextual distinctions. Similar behavior has been observed in prior studies on heterophilous or noisy graphs, where deep aggregation amplifies irrelevant neighbor information and degrades representation quality \citep{Velickovic2018GAT, Rusch2023OversmoothingSurvey, Zhu2020BeyondHomophily}.
Overall, these findings indicate that small, homogeneous graphs benefit from moderate propagation depth, whereas large and heterogeneous graphs require shallower aggregation to prevent semantic noise accumulation. We therefore adopt the two-layer configuration as the default, providing a strong balance between contextual expressiveness and stability across datasets.\\
\textbf{Zero-Shot Cross-Modal Retrieval Evaluation.}
To evaluate the semantic robustness and modality independence of our Text2Graph VPR framework, we conduct zero-shot cross-modal retrieval experiments. In these tests, the model is applied to a modality it was not trained on, without any additional fine-tuning. We selected 10\% of the Amman query images and 10\% of the database images and asked several people to describe these scenes using a structured instruction prompt indicating which objects to identify and how spatial relationships should be expressed. The resulting human-written descriptions were manually reviewed for clarity and consistency before being parsed into scene graphs.\\
In the first setting (Experiment A), human-written descriptions were used as queries, while the database remained image-derived. This represents zero-shot language-only localization, where the model receives no visual input at test time. Success in this setting indicates that the learned representations capture semantic and structural properties of places rather than relying on dataset-specific linguistic patterns. Furthermore, to assess domain generalization, we perform a complementary experiment (Experiment B), in which the database descriptions are replaced with human-written descriptions, while the queries originate from image-derived descriptions. This second setup tests whether the learned graph embeddings remain stable across human- and model-generated language domains. These two configurations, summarized in Table 9 as Zero-shot cross-modal retrieval (Experiment A) and Cross-domain text alignment (Experiment B), evaluate the framework’s ability to generalize across both modality (image–text) and linguistic domain (model-generated vs. human-written). \\
\begin{table}[t]
\caption{Zero-shot retrieval using human-written queries and human-written database descriptions (test on Amman).}
\centering
\resizebox{\textwidth}{!}{%
\small
\begin{tabular}{|l|ccc|}
\hline
Experiment & Query Source & Database Source & Amman \\
           &              &                 & 1 / 5 / 20 \\
\hline
Normal Retrieval (Baseline) &
Image-derived &
Image-derived &
22.6 / 51.4 / 83.8 \\

Experiment A --- Zero-Shot Cross-Modal Retrieval &
Human-written &
Image-derived &
22.6 / 54.2 / 83.5 \\

Experiment B --- Cross-Domain Text Alignment &
Image-derived &
Human-written &
21.4 / 50.2 / 80.7 \\
\hline
\end{tabular}
}
\label{tab:zeroshot_amman}
\end{table}

As shown in Table 9, the proposed framework maintains strong retrieval performance in both cases without retraining, confirming its zero-shot and modality-independent properties. The zero-shot capability of our method stems from representing each description as a scene graph composed of objects and their pairwise relations, rather than directly relying on the raw language form. Because the model operates on structured semantic elements—nodes and edges—it remains robust to variability in sentence structure, writing style, or vocabulary in human-written descriptions. Even when humans introduce synonyms, omit details, or write freely, the underlying graph remains consistent with the key objects and spatial or relational cues extracted from the scene. Unlike CLIP-style approaches that rely on pixel–text alignment, our framework operates entirely in the semantic-graph domain, enabling interpretable and language-grounded localization even in the absence of visual input. \\
\section{Discussion}
\subsection{Advantages of Using Textual Descriptions for Place Recognition}
\begin{itemize}
   \item Robustness to Environmental Variations: Text abstracts away low-level pixel details and emphasizes high-level semantic content. As a result, descriptions remain relatively stable across varying lighting conditions, seasonal changes, dynamic objects, and occlusions, enhancing place recognition robustness in challenging real-world scenarios.
   \item Enhanced Generalization Across Domains: By encoding relational semantics, the model captures structural regularities that remain valid across domains and environmental conditions. This enables generalization between geographically or temporally distinct environments, as demonstrated in our cross-city experiments (Section 5.3).
   \item Zero-Shot and Modality-Independent Retrieval: A key strength of the Text2Graph framework is its ability to operate in a zero-shot cross-modal setting—retrieving locations from human-written text without any retraining on that modality. Because both visual and linguistic inputs are converted into the same semantic-graph representation, the model can localize scenes even when no visual query is available. This property demonstrates deep semantic alignment beyond pixel-level correlation, setting it apart from traditional image-text models such as CLIP.
   \item Efficiency in Storage and Retrieval: Graph-based embeddings derived from compact textual inputs require significantly less storage and bandwidth than dense visual features and enabling faster retrieval. Moreover, similarity computation in the graph-semantic space is efficient and less sensitive to visual redundancy.
   \item Interpretability and Human-Readable Representations: A core advantage of the Text2Graph framework is its interpretability. Both intermediate textual descriptions and final graph representations are human-readable, enabling intuitive debugging, error analysis, and qualitative reasoning. This aligns with the goals of expert systems by providing transparent, semantically grounded decision evidence. This transparency contrasts with black-box visual embeddings, making the model’s decisions explainable and verifiable—a key property for real-world deployment in autonomous or safety-critical systems.
   \item Interoperability and Knowledge Integration: Textual and graph-based representations can easily interact with external knowledge sources, such as ontologies, map databases, or large language models (LLMs). This opens the door to higher-level reasoning and cross-task integration—e.g., connecting localization with route description, dialogue-based navigation, or question answering. Such interoperability reflects the broader vision of knowledge-driven intelligent systems.
\end{itemize}
Thus, while current accuracy may trail purely visual baselines, the proposed representation provides a more extensible and semantically grounded foundation for future multimodal localization systems.\\
\subsection{Limitations}
\begin{itemize}
     \item	Semantic Ambiguity in Outdoor Environments: Urban outdoor scenes often exhibit repetitive structures (roads, vegetation, fences), leading to similar textual and graph representations for distinct places. This semantic redundancy limits the discriminative power of purely semantic features, especially in visually homogeneous regions. While our dual-similarity framework (GAT-based embeddings + SP kernel) alleviates part of this issue by combining structural and relational cues, ambiguity remains a fundamental challenge for text-based VPR.
     \item Performance Gap with State-of-the-Art: While the proposed method performs competitively under appearance changes, it currently falls short of the top-performing image-based systems in terms of raw accuracy. This gap is likely due to information loss during the conversion from images to text and graphs. Addressing this will require richer captioning models, improved graph consistency, and tighter alignment between textual semantics and spatial grounding.
     \item Dependency on description Quality: The overall system performance heavily depends on the quality and completeness of the textual descriptions generated by the vision-language model. Inaccurate or overly generic descriptions may lead to suboptimal or ambiguous graph representations, negatively impacting retrieval.
     \item Parsing and Graph Construction Challenges: Translating free-form text into structured scene graphs can introduce parsing errors, misidentified relationships, or inconsistent object representations. Moreover, aligning and merging graphs from multiple frames or sequences remains a non-trivial challenge that can affect graph similarity calculations. Developing more reliable, constraint-guided parsing is an important direction for improving robustness.
\end{itemize}
\subsection{Cross-City and Cross-Condition Generalization}
The Oxford RobotCar results primarily assess robustness to environmental appearance changes within the same city, such as seasonal or illumination variations. In contrast, the MSLS benchmark evaluates a much harder setting—cross-city generalization—where the model is trained in one city (Melbourne) and tested on entirely different cities (Amman and San Francisco) featuring distinct architectural styles, lighting conditions, and urban layouts.\\
Traditional visual methods (e.g., NetVLAD \citep{Arandjelovic2016NetVLAD} and AP-GeM \citep{Revaud2019APLoss}) achieve high accuracy ($>$90\%) under same-city conditions but degrade sharply on unseen cities ($<$60\%) due to overreliance on visual texture and color cues. Our Text2Graph VPR, on the other hand, represents scenes through semantic and structural relations, which remain invariant across cities.
Although trained only on a simplified subset of Melbourne, our model maintains competitive or superior performance on unseen MSLS cities, demonstrating that semantic-structural abstraction enables robust transfer across diverse urban domains. This property is particularly relevant for intelligent decision-support systems, where the ability to generalize beyond training conditions is essential for reliability and deployment.
\section{Future Directions}
Future work should focus on improving scene graph construction to enhance semantic richness and consistency, and on developing multimodal fusion strategies that combine textual and visual cues. Large vision–language models (e.g., LVLM-VPR \citep{Wang2024LVLMVPR}, MSSPlace \citep{Melekhin2025MSSPlace}) suggest promising directions for integrating cross-modal features. Furthermore, scaling the framework toward end-to-end training—optimizing visual encoding, captioning, and graph embedding jointly—may yield more cohesive and higher-performing systems. Another promising direction is incorporating explicit reasoning modules or rule-based consistency checks, enabling Text2Graph VPR to evolve into a more complete semantic expert system for localization
\section{Conclusion}
In this work, we introduced Text2Graph VPR, a novel framework that performs visual place recognition through semantic scene graphs derived from vision–language models. By generating textual descriptions for each frame, parsing them into structured graphs, and fusing semantic embedding and structural similarity, our approach enables interpretable and modality-independent localization. The emphasis on human-readable intermediate representations aligns with expert-system principles and provides transparent evidence for each retrieval decision. Through extensive experiments on Oxford RobotCar \citep{Maddern2017RobotCar}, Boreas, and MSLS \citep{Warburg2020MSLS}, we demonstrated that text-based representations can achieve competitive retrieval performance, particularly under challenging viewpoint and illumination changes where visual-only systems often degrade. Our analysis further showed that adaptive dual-similarity fusion plays a crucial role in balancing scene-level semantics and graph structure. While the current method does not yet match the accuracy of state-of-the-art visual baselines, it establishes a transparent and extensible foundation for multimodal VPR. Future work will explore joint graph refinement, large-scale language model alignment, and the integration of cross-modal supervision to further strengthen semantic localization. Ultimately, the proposed approach supports the development of interpretable, knowledge-aware localization systems suitable for intelligent autonomous platforms.
\section*{CRediT authorship contribution statement}

\textbf{Saeideh Yousefzadeh:} Conceptualization, Methodology, Software, Formal analysis,
Investigation, Writing – original draft, Writing – review \& editing.

\textbf{Hamidreza Pourreza:} Methodology, Software, Validation, Writing – review \& editing.
\section*{Declaration of Competing Interest}

The authors declare that they have no known competing financial interests or personal
relationships that could have appeared to influence the work reported in this paper.
\section*{Data availability}

The datasets used in this study are publicly available. Oxford RobotCar is available at
\url{https://robotcar-dataset.robots.ox.ac.uk/}, and MSLS is available at
\url{https://www.mapillary.com/dataset/places}.
\section*{Declaration of generative AI and AI-assisted technologies in the manuscript preparation process}
During the preparation of this manuscript, the authors used ChatGPT solely to assist with language editing and improving the clarity and fluency of the text. 
All content generated or suggested by the AI tool was carefully reviewed, verified, and revised by the authors, who take full responsibility for the scientific content, interpretations, and conclusions presented in this paper.

\bibliography{references_eswa}

\end{document}